\documentclass{article}
\usepackage[a4paper, total={7.2in, 10.5in}]{geometry}
\usepackage{graphicx} 
\usepackage{amsmath,amsthm}
\usepackage{amssymb}
\usepackage[english]{babel}
\usepackage{latexsym}
\usepackage{amsfonts}
\usepackage{graphicx}
\usepackage{float}
\usepackage{graphics}
\usepackage{epsfig}
\usepackage{url}
\usepackage{array}
\usepackage{ragged2e}
\usepackage{multirow}
\usepackage{amssymb} 
\usepackage{hyperref}
\hypersetup{
    colorlinks=true,
    linkcolor=blue,
    filecolor=blue,      
    urlcolor=blue,
}
\usepackage{subcaption}
\usepackage{csquotes}

\usepackage[backend=biber, style=numeric, sorting=none]{biblatex}
\AtBeginBibliography{\small}

\small {\title{Cell Phone Image-Based Persian Rice Detection and Classification Using Deep Learning Techniques}}
\author{Mahmood Saeedi Kelishami\thanks{Department of Mathematics, Rasht Branch, Islamic Azad University, Email: \href{mailto:mskelishami@gmail.com}{mskelishami@gmail.com}} \and Amin Saeidi Kelishami\thanks{Department of Computer Engineering, Sharif University of Technology, Email: \href{mailto:amin.saeidi.1997@gmail.com}{amin.saeidi.1997@gmail.com}} \and Sajjad Saeedi Kelishami\thanks{Department of Architecture, Rasht Branch, Islamic Azad University, Email: \href{mailto:sajjadsaeedi2018@gmail.com}{sajjadsaeedi2018@gmail.com}}} 

\date{April 2024}

\begin{document}

\maketitle

\begin{abstract}
This study introduces an innovative approach to classifying various types of Persian rice using image-based deep learning techniques, highlighting the practical application of everyday technology in food categorization. Recognizing the diversity of Persian rice and its culinary significance, we leveraged the capabilities of convolutional neural networks (CNNs), specifically by fine-tuning a ResNet model for accurate identification of different rice varieties and employing a U-Net architecture for precise segmentation of rice grains in bulk images. This dual-methodology framework allows for both individual grain classification and comprehensive analysis of bulk rice samples, addressing two crucial aspects of rice quality assessment. Utilizing images captured with consumer-grade cell phones reflects a realistic scenario in which individuals can leverage this technology for assistance with grocery shopping and meal preparation. The dataset, comprising various rice types photographed under natural conditions without professional lighting or equipment, presents a challenging yet practical classification problem. Our findings demonstrate the feasibility of using non-professional images for food classification and the potential of deep learning models, like ResNet and U-Net, to adapt to the nuances of everyday objects and textures. This study contributes to the field by providing insights into the applicability of image-based deep learning in daily life, specifically for enhancing consumer experiences and knowledge in food selection. Furthermore, it opens avenues for extending this approach to other food categories and practical applications, emphasizing the role of accessible technology in bridging the gap between sophisticated computational methods and everyday tasks.
\end{abstract}


\section{INTRODUCTION}

Rice stands as a foundational agricultural product and staple food, instrumental in feeding more than half of the global population. It is a significant source of sustenance for approximately 3.5 billion individuals worldwide and represents a crucial element of food security, with an annual production surpassing 500 million tons. Beyond its role as a dietary staple, rice cultivation is a vital economic activity, offering substantial income for countless farmers across various regions. The emphasis on sophisticated and accurate methodologies for rice quality and classification has become increasingly prominent. This urgency is driven by the potential to enhance market acceptability, minimize rejection rates, and elevate the economic gains for producers through reliable quality assurance practices \cite{razavi2024resnet}.

In the realm of agricultural quality assessment, traditional methods often depend on manual inspection based on visual appearance and smell, which, despite their widespread use, suffer from limitations in speed, accuracy, and reliability, particularly for those without extensive experience. Recent advancements in technology have paved the way for the application of data mining and machine learning techniques, marking a significant leap in enhancing the efficiency and precision of rice classification processes. These innovative approaches utilize detailed feature extraction from images, analyzing color, shape, and textural characteristics to differentiate rice varieties and ascertain their quality with unprecedented accuracy \cite{cinar2019classification, majumdar2000classification, visen2004comparison, wan2017novel, silva2013classification}.

Sumaryanti et al. present a system designed for the identification of rice varieties using image processing techniques and a LVQ neural network algorithm. The system analyzes rice images, extracting features such as color, morphology, and texture. By combining these features, the model achieves an average accuracy of 70.3 percent, with the highest classification accuracy of 96.6 percent for the Mentik Wangi rice variety and the lowest accuracy of 30 percent for Cilosari \cite{sumaryanti2015digital}. 

However, despite the critical importance of rice classification and the promising results of preliminary research, the area remains relatively underexplored, with a limited number of studies addressing the full spectrum of possibilities offered by modern computational methods. For example, Mavaddati introduced an approach using sparse representation and dictionary learning to enhance rice quality classification, showing significant potential for these techniques to surpass traditional machine learning models in terms of accuracy and reliability \cite{mavaddati2018rice}. , Qadria et al. aimed to assess machine vision (MV) techniques for classifying six Asian rice varieties: Kachi-Kainat, Kachi-Toota, Kainat-Pakki, Super-Basmati-Kachi, Super-Basmati-Pakki, and Super-Maryam-Kainat. These varieties are commonly cultivated in Pakistan, China, India, Bangladesh, and neighboring countries. The authors captured digital images of rice grains using a cell phone camera in an open climate. From these images, they extracted binary, histogram, and texture features. Leveraging feature optimization, they deployed five MV classifiers—LMT Tree (LMT-T), Meta Classifier via Regression (MCR), Meta Bagging (MB), Tree J48 (T-J48), and Meta Attribute Select Classifier (MAS-C), resulted in the highest overall accuracy was observed with the LMT-Tree model \cite{qadri2021machine}.

Koklu et al. explore the application of deep learning techniques for the classification of rice varieties. They utilize three different models: Artificial Neural Network (ANN), Deep Neural Network (DNN), and Convolutional Neural Network (CNN). The authors create models based on both feature datasets (including morphological, shape, and color features) and image datasets. Their findings demonstrate the successful application of deep learning in rice variety classification, with potential implications for agriculture and food security \cite{koklu2021classification}. In a comprehensive study,  Ahad et al. conducted a comprehensive comparison of six CNN-based deep learning architectures—DenseNet121, Inceptionv3, MobileNetV2, resNext101, Resnet152V, and Seresnext101—using a database of nine of the most epidemic rice diseases in Bangladesh. By analyzing the performance of these architectures, the authors aimed to enhance the accuracy of rice disease classification \cite{ahad2023comparison}. 

Razavi et al. explore the application of ResNet deep models and transfer learning techniques for the classification and quality assessment of rice cultivars. Leveraging the power of convolutional layers, max-pooling, and fully connected layers, ResNet effectively learns robust representations from rice images. By combining these techniques, the authors achieve impressive accuracy in identifying different rice varieties based on visual features \cite{razavi2024resnet}. Din et al. address the critical task of rice grain variety classification using a computer vision system called RiceNet. This system is contingent upon a Deep Convolutional Neural Network (DCNN) framework, which effectively identifies five unique groups of rice grain varieties. Leveraging pre-trained architectures such as InceptionV3 and InceptionResNetV2 \cite{din2024ricenet}. Farahnakian et al. delve into the critical task of rice grain classification using novel deep learning models. Traditional machine learning approaches often require intricate feature engineering, but the authors explore the effectiveness of contemporary deep learning architectures. They evaluate models such as Residual Network (ResNet) \cite{he2016deep}, Visual Geometry Group (VGG) network \cite{simonyan2014very}, EfficientNet \cite{tan2019efficientnet}, and MobileNet \cite{howard2017mobilenets} on a dataset containing 75,000 rice images across five different rice categories \cite{farahnakian2024comparative}.

On the other hand for rice image segmentation, Tan et al. present an algorithm designed to address the challenge of segmenting and counting touching hybrid rice grains. The goal is to enable automatic evaluation of seeding performance. The proposed algorithm combines several techniques, including the watershed algorithm, an improved corner point detection algorithm, and a neural network classification algorithm. To mitigate over-segmentation caused by the watershed algorithm, the authors apply wavelet transform and Gaussian filter to enhance grayscale image contrast and reduce noise. The algorithm accurately merges over-segmented regions by detecting whether splitting lines’ end points coincide with corner points.\cite{tan2019segmentation}. Nagoda et al. propose an innovative approach for segmenting and classifying rice samples based on color and texture features. Leveraging image processing techniques, they acquire rice images using a CCD camera and apply grayscale conversion, noise reduction, and morphological operations. The watershed algorithm effectively handles touching and overlapping rice kernels. Extracted Local Binary Pattern (LBP) texture features and color information are then used to predict rice sample objects using a Linear Kernel-based Support Vector Machine (SVM) \cite{nagoda2018rice}.

Building upon these advancements, our study introduces a novel approach by utilizing cell phone images for the classification of Persian rice varieties. This innovation emphasizes the practicality of employing everyday technology in agricultural and food quality applications, making sophisticated classification methods accessible to a broader audience. By leveraging the ubiquity of cell phones, our research demonstrates how deep learning models can be adapted for use in everyday settings, providing consumers and farmers with a powerful tool for identifying rice varieties and assessing quality with ease. This approach not only opens new pathways for agricultural technology but also underscores the potential of integrating machine learning into daily life for enhanced food security and economic efficiency.

\section{PROBLEM DEFINITION}

The central challenge addressed by our study revolves around two pivotal aspects of rice classification: single grain classification and bulk prediction. These twin facets underscore the complexity and multidimensional nature of rice quality assessment and variety identification, critical for both agricultural production and consumer choice.

\textbf{Single Grain Classification:} The first aspect of the problem lies in accurately classifying individual rice grains captured in an image into one of seven distinct varieties: Ali Kazemi, Anbar Boo (Ghareeb), Hashemi, Khazar, Sadree Dom Siahe, Sadree Dom Zard, and Shirodi. This level of classification demands a nuanced understanding of the subtle variations between rice types, which are often indistinguishable to the unaided eye. Traditional methods fall short in this regard, due to their reliance on subjective assessment and the inherent variability in human judgment. The need for an automated, accurate, and reliable system to perform such detailed classification is evident, as it would significantly enhance the precision of rice quality control and variety identification.

\textbf{Bulk Prediction:} The second aspect focuses on the ability to analyze images containing a mixture of rice grains and estimate the percentage composition of each variety present. This task involves not only the identification of individual grains within a bulk sample but also the segmentation and quantification of each type, presenting a unique set of challenges. Such capability would be invaluable for assessing the quality and composition of rice in bulk purchases or shipments, providing a comprehensive overview of the product mix. Current approaches, typically manual or semi-automated, are labor-intensive, time-consuming, and prone to inaccuracies, highlighting the necessity for an advanced solution that can streamline the process and deliver precise, reliable results.

The overarching problem, therefore, encompasses the development of an efficient, accessible, and accurate methodology for both single grain classification and bulk rice prediction. Addressing these challenges requires leveraging the latest advancements in machine learning and image processing, alongside the ubiquitous technology of cell phone cameras, to democratize the classification process. By doing so, our research aims to transform rice quality assessment from a cumbersome and error-prone task into a streamlined, user-friendly, and precise operation, opening new avenues for technology adoption in agriculture and retail, and fostering greater food security and market transparency.

\section{METHODOLOGY}
\begin{figure}[!ht]
    \centering
    \includegraphics[scale=0.5]{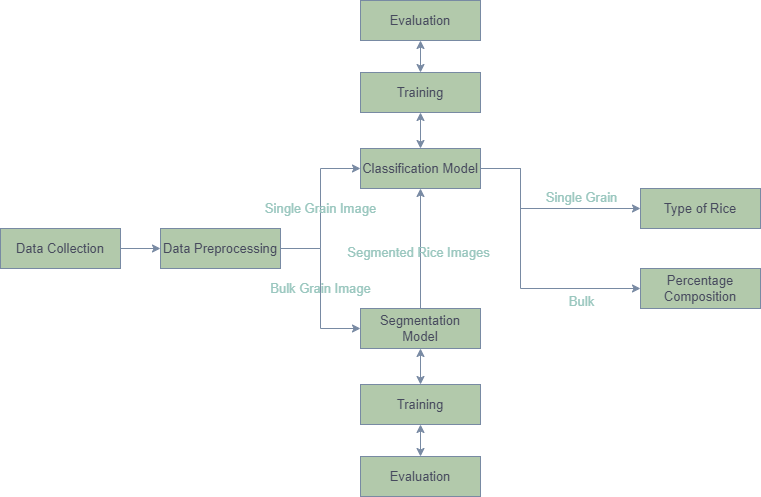}
    \caption{Schematic representation of the dual-model framework employing ResNet for single grain classification and U-Net for segmentation in bulk prediction tasks. This diagram illustrates the process flow from data collection and preprocessing through to model training and evaluation.}
    \label{fig:MainFramework}
\end{figure}

This research employs a comprehensive methodology, as depicted in Fig.\ref{fig:MainFramework},  combining state-of-the-art deep learning techniques to tackle the challenges of Persian rice classification and bulk prediction. Our approach is underpinned by two main objectives: accurately classifying individual rice grains into seven distinct varieties and estimating the percentage composition of these varieties in bulk samples. To achieve these goals, we implemented a dual-model framework utilizing ResNet (ResNet50) for classification and U-Net \cite{ronneberger2015u} for segmentation task. For the task of classifying individual rice grains, we fine-tuned a pre-existing ResNet model, chosen for its deep residual learning framework that enables the training of deeper networks by addressing the vanishing gradient problem. This aspect is crucial for learning the intricate patterns and subtle differences between rice varieties. The model was trained on cropped images of single rice grains, with the objective of minimizing classification error across the seven target categories. The bulk prediction task required the segmentation of mixed rice grain images to identify and quantify each grain type present. We utilized the U-Net architecture, renowned for its effectiveness in medical image segmentation, adapted to our context for segmenting rice grains within bulk images. This model employs a contracting path to capture context and a symmetric expanding path that enables precise localization, making it ideal for segmenting closely packed or overlapping rice grains. 

Both models were trained using a split of the dataset into training, validation, and test sets, ensuring a comprehensive evaluation of their performance. The training process involved adjusting hyperparameters such as learning rate, batch size, and number of epochs to optimize model accuracy. The models were implemented using Python, TensorFlow (for U-net), and PyTorch (for ResNet) with experiments run on GPU-accelerated hardware to expedite the training process. Data augmentation techniques like rotation, scaling, and flipping were applied to enrich the dataset and improve model generalizability.

\section{EXPERIMENTS}
To assess the effectiveness of our dual-model framework in addressing the challenges of Persian rice classification and bulk prediction, a series of experiments were designed and executed. These experiments were aimed at not only validating the accuracy and efficiency of our proposed methodology but also at exploring its practical applicability in real-world scenarios.

\subsection{Experimental Settings}

The experimental settings are meticulously devised to ensure the robust evaluation of our models across varied conditions. This includes detailed consideration of dataset characteristics, image acquisition methods, and the specific configurations of our deep learning models. The settings are structured to replicate practical conditions as closely as possible, thereby enhancing the relevance and applicability of our findings.

\subsubsection{Datasets}

\begin{figure}[!hb]
    \centering
    \begin{subfigure}[b]{0.3\linewidth}
        \includegraphics[width=\linewidth]{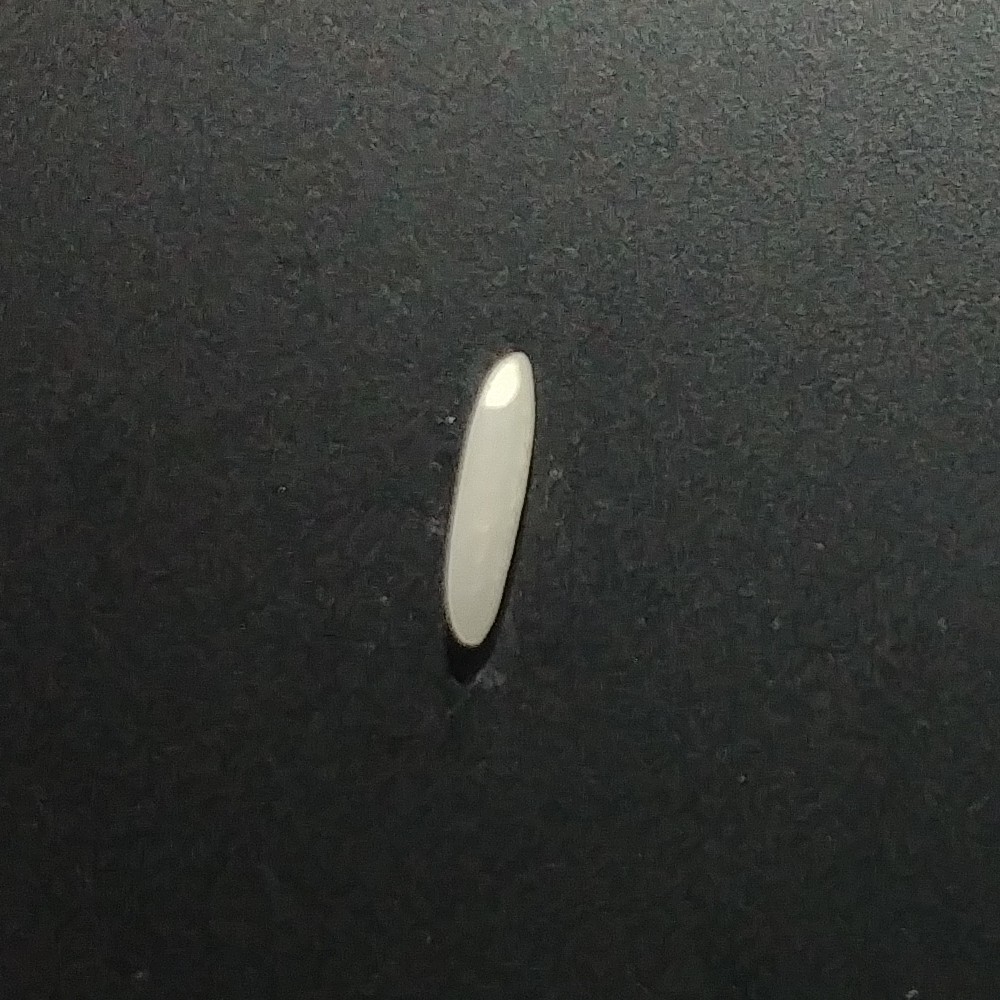}
        \caption{Ali Kazemi}
    \end{subfigure}
    \begin{subfigure}[b]{0.3\linewidth}
        \includegraphics[width=\linewidth]{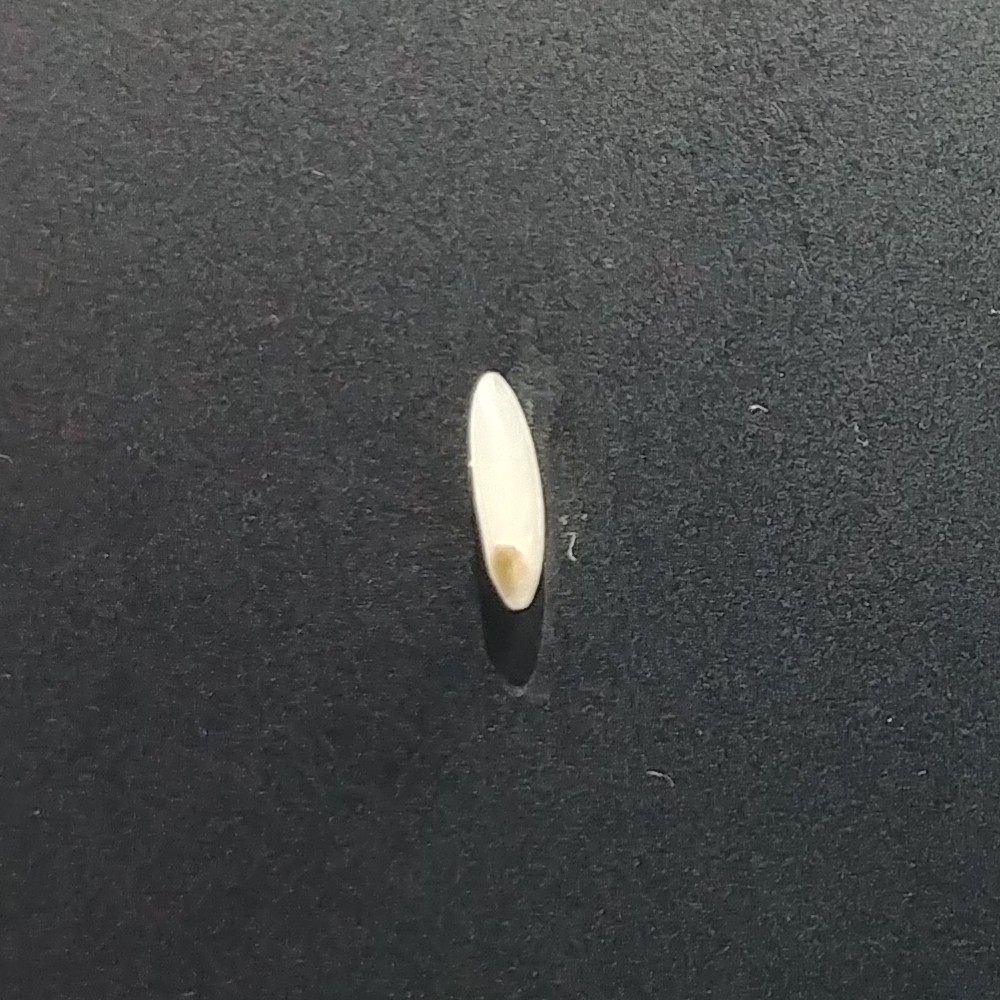}
        \caption{Anbar Boo (Ghareeb)}
    \end{subfigure}
    \begin{subfigure}[b]{0.3\linewidth}
        \includegraphics[width=\linewidth]{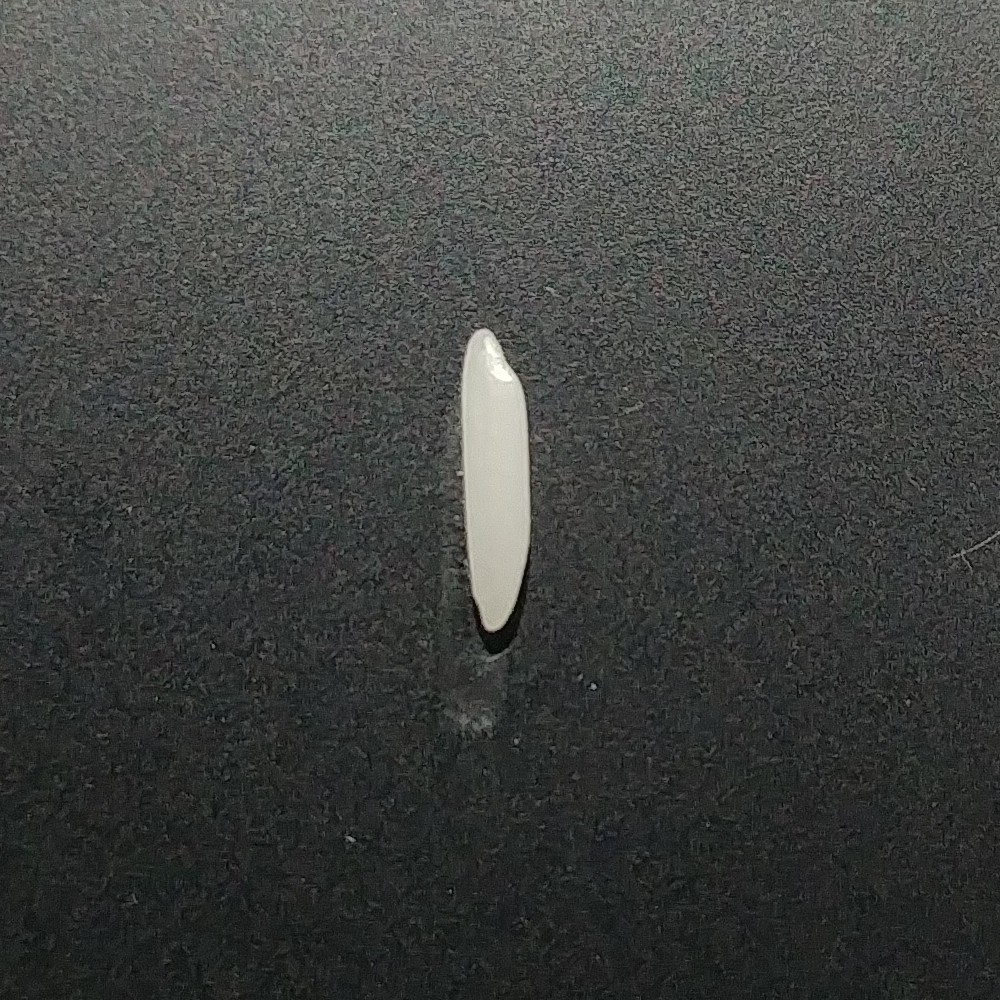}
        \caption{Sadree Dom Siah}
    \end{subfigure}
    
    \begin{subfigure}[b]{0.3\linewidth}
        \includegraphics[width=\linewidth]{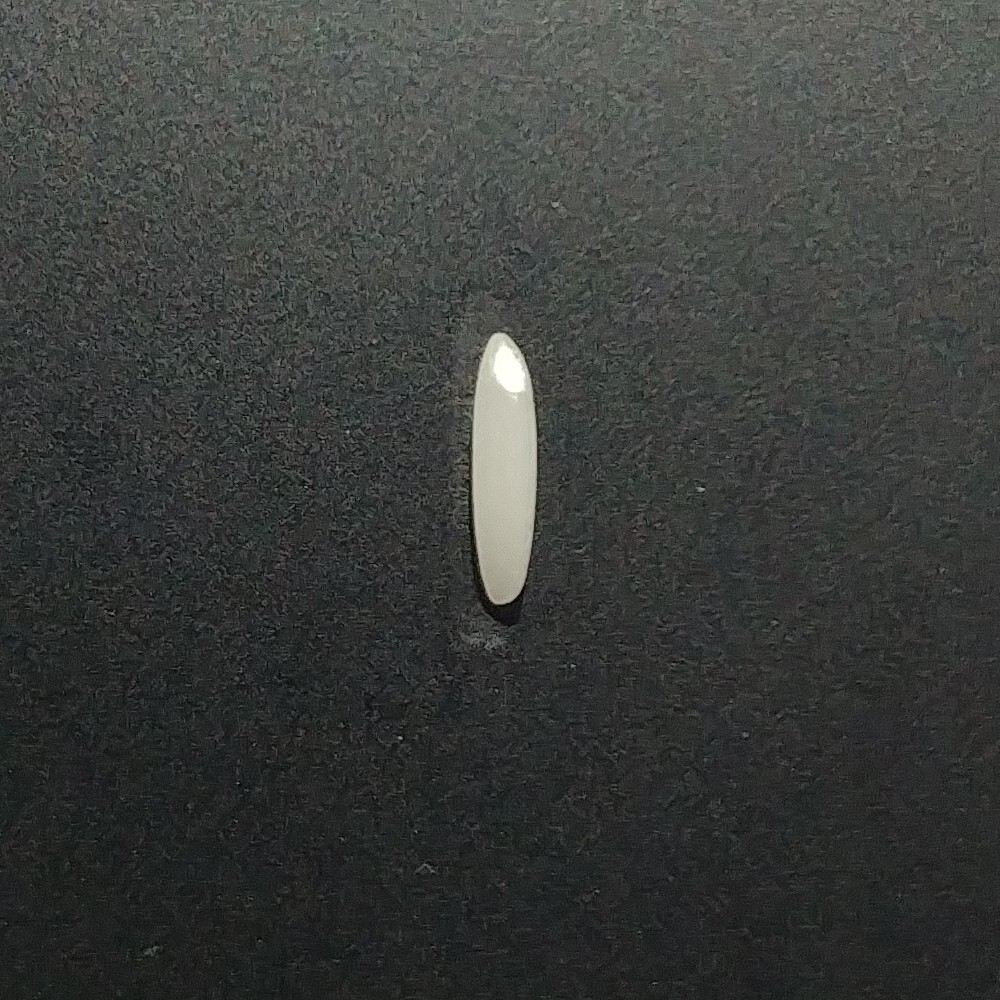}
        \caption{Sadree Dom Zard}
    \end{subfigure}
    \begin{subfigure}[b]{0.3\linewidth}
        \includegraphics[width=\linewidth]{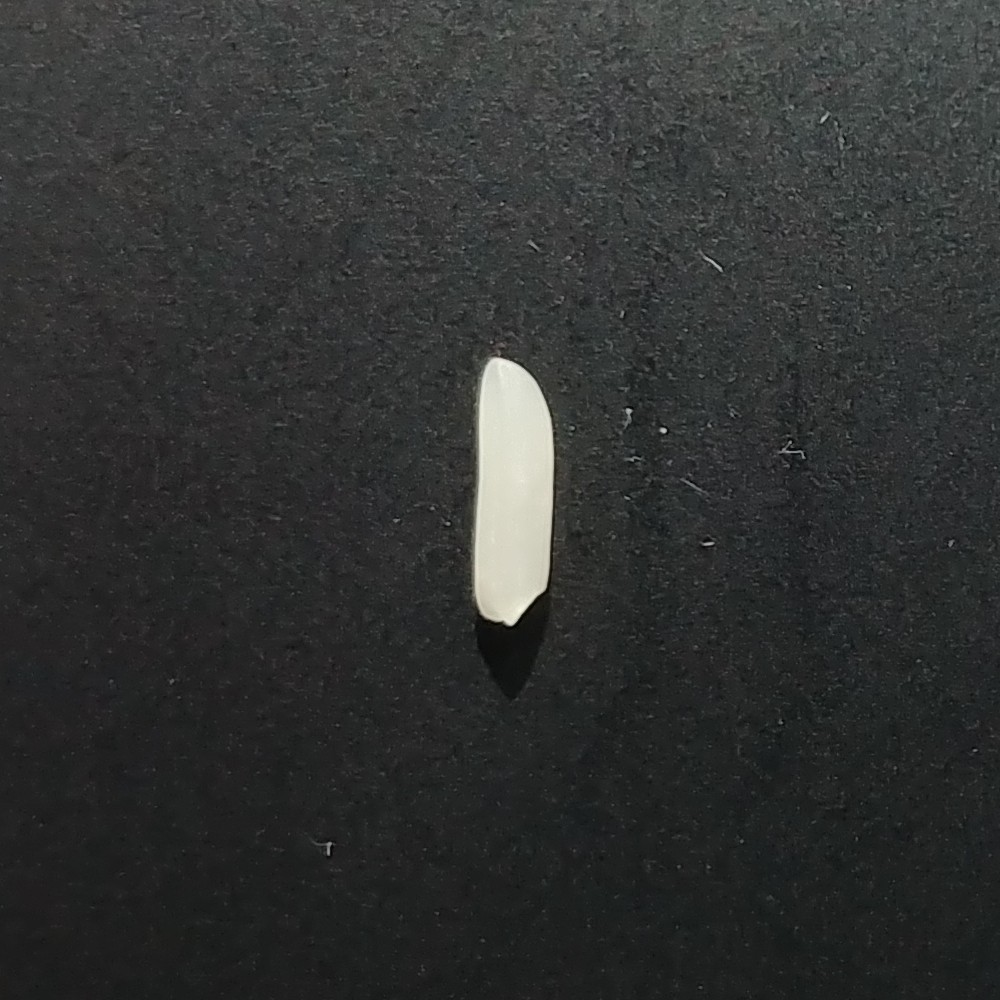}
        \caption{Hashemi}
    \end{subfigure}
    \begin{subfigure}[b]{0.3\linewidth}
        \includegraphics[width=\linewidth]{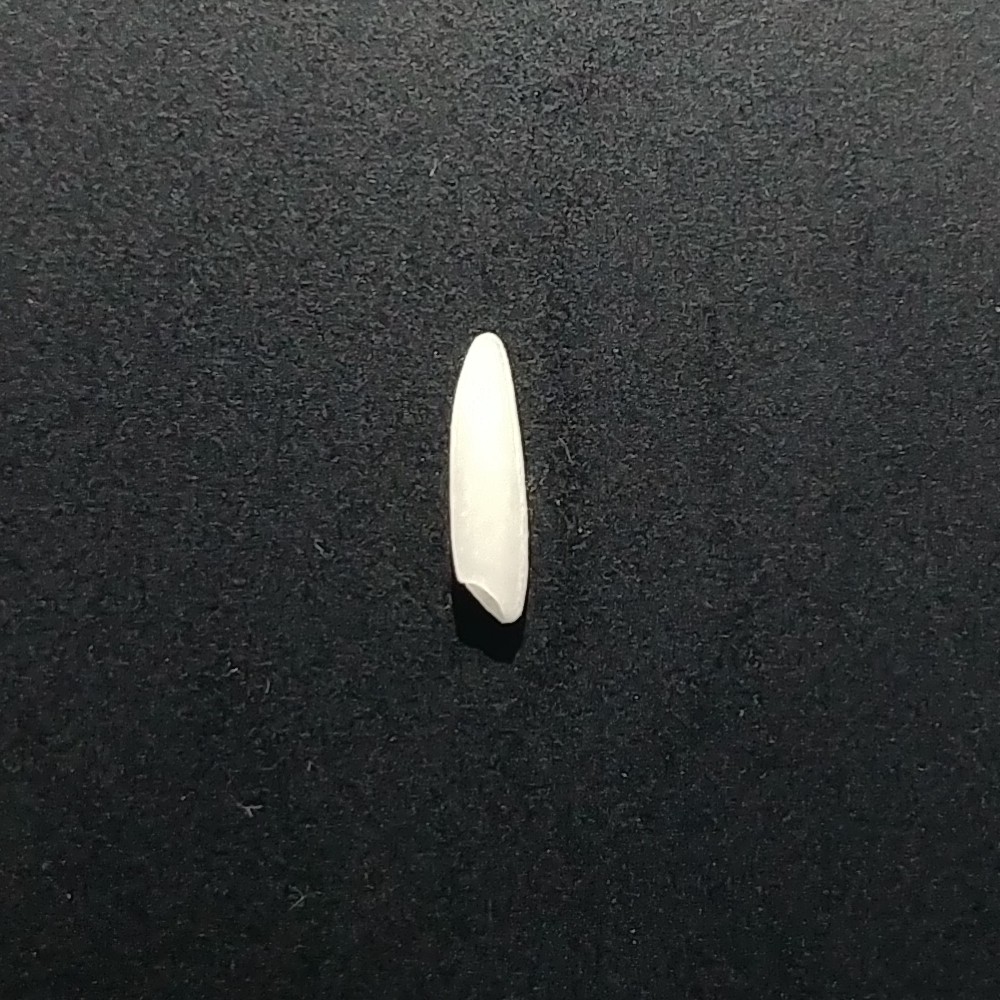}
        \caption{Khazar}
    \end{subfigure}
    
    \begin{subfigure}[b]{0.3\linewidth}
        \centering
        \includegraphics[width=\linewidth]{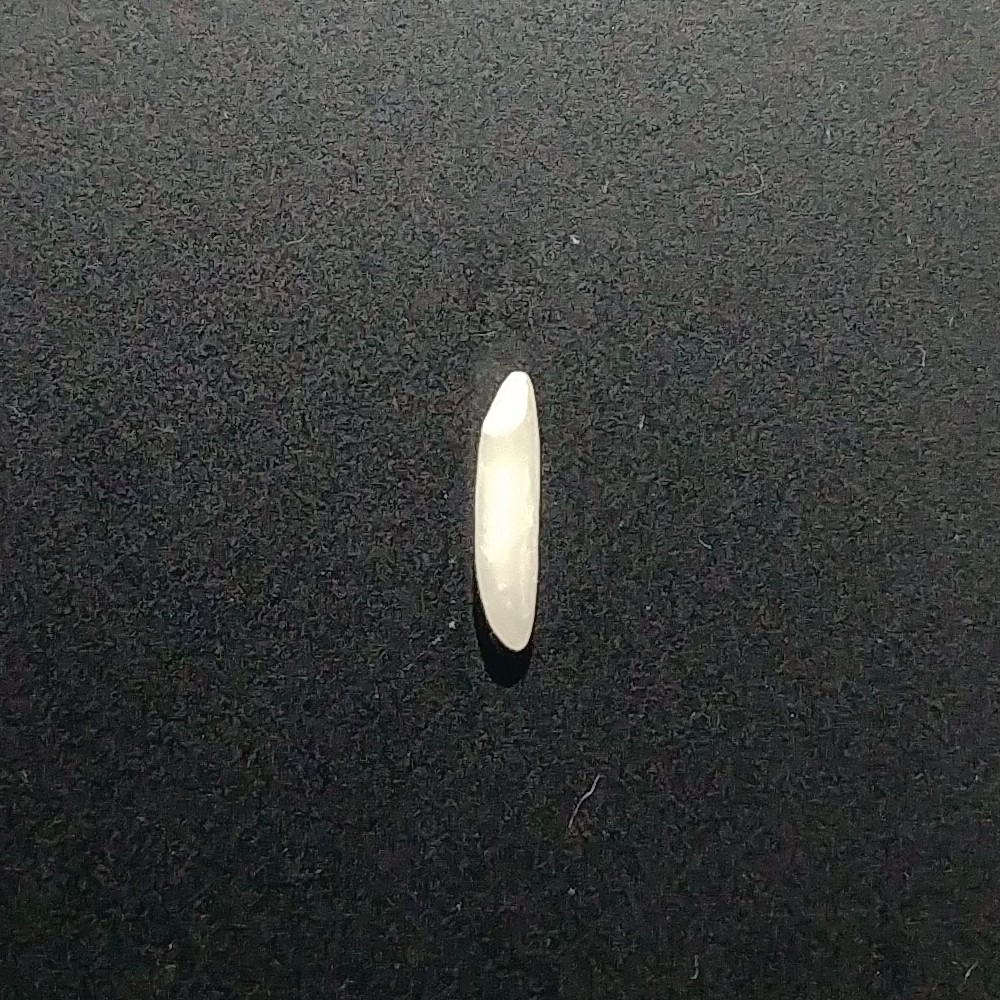}
        \caption{Shirodi}
    \end{subfigure}
    
    \caption{Visual representation of the seven Persian rice varieties classified in this study, showcasing the distinctive appearance of each variety. These images serve as a basis for both single grain classification and bulk rice prediction tasks, demonstrating the diversity and complexity of rice types found in Persian cuisine.}
    \label{fig:SingleRice}
\end{figure}

\begin{figure}[!ht]
    \centering
    \begin{subfigure}{0.48\linewidth}
        \includegraphics[width=\linewidth]{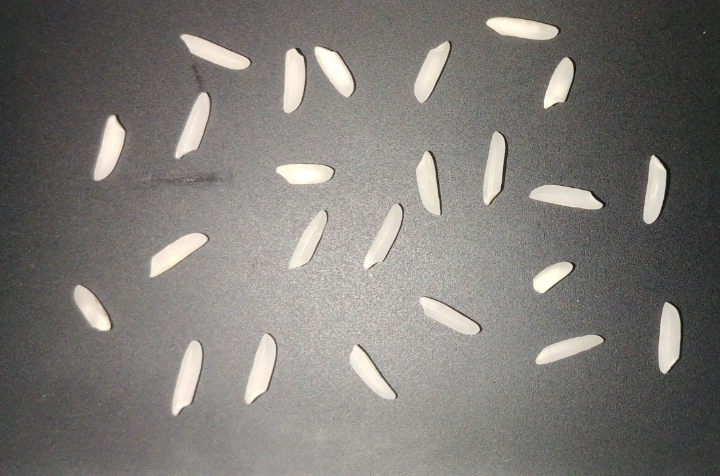}
        \caption{Bulk Rice Sample}
    \end{subfigure}
    \hfill
    \begin{subfigure}{0.48\linewidth}
        \includegraphics[width=\linewidth]{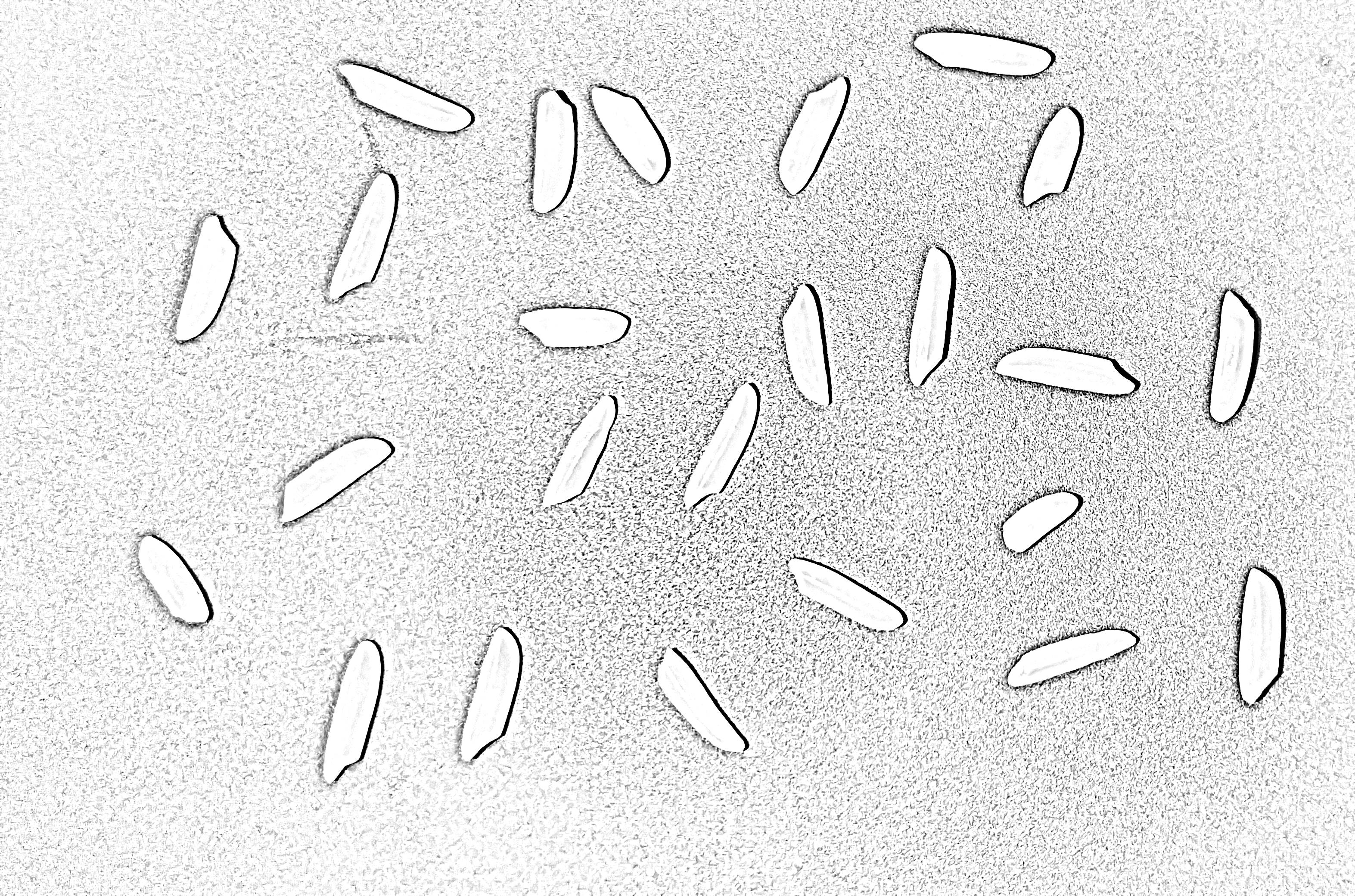}
        \caption{Segmentation Mask for Bulk Rice}
    \end{subfigure}
    \caption{Comparison of a bulk rice sample with its corresponding segmentation mask. The left image showcases a diverse mixture of rice grains, while the right image illustrates the sample segmentation mask utilized and generated by the U-Net model, highlighting the model's ability to identify and segment individual grains within the bulk sample. This process is critical for estimating the percentage composition of different rice varieties in mixed samples.}
    \label{fig:BulkRice}
\end{figure}

Our study utilizes two datasets. The first, referred to as the single grain dataset, contains 550 different images categorized into 7 types of Persian rice. Each image was captured using two types of cell phones: a LG G6 and a Xiaomi Redmi Note 12, as depicted in Fig.\ref{fig:SingleRice}. This diversity in image capture devices is intended to simulate the variability encountered in real-life applications. The second dataset comprises 100 pictures of bulk rice images, each accompanied by original images and their segmentation masks, illustrated in Fig.\ref{fig:BulkRice}. This dataset is crucial for testing the efficacy of our segmentation approach in handling complex, mixed rice samples.

\subsubsection{Evaluation Metrics}

For the evaluation of our models, we employ distinct metrics tailored to the specific tasks of classification and segmentation. These metrics were selected for their ability to provide a comprehensive overview of model performance, encompassing aspects of accuracy, precision, recall, and the quality of segmentation.

\begin{itemize}
    \item \textbf{Classification Metrics:}
\begin{itemize}
    \item \textbf{Accuracy:} Defined as the ratio of correctly predicted observations to the total observations. 
    \[Accuracy = \frac{TP+TN}{TP+TN+FP+FN}\]
    \item \textbf{Precision (P):} The ratio of correctly predicted positive observations to the total predicted positive observations.
    \[Precision = \frac{TP}{TP+FP}\]
    \item \textbf{Recall (R):} The ratio of correctly predicted positive observations to all observations in the actual class.
    \[Recall = \frac{TP}{TP+FN}\]
    \item \textbf{F1 Score:} The weighted average of Precision and Recall.
    \[F1 = 2 \times \frac{Precision \times Recall}{Precision + Recall}\]
\end{itemize}

\item \textbf{Segmentation Metrics:}
\begin{itemize}
    \item \textbf{Intersection over Union (IoU):} Also known as the Jaccard index, measures the overlap between two boundaries.
    \[IoU = \frac{Area \ of \ Overlap}{Area \ of \ Union}\]
\end{itemize}

\end{itemize}

\subsection{Results}
The results from our experiments on Persian rice classification and segmentation present significant insights into the capabilities and challenges of employing deep learning models for agricultural product classification. Through meticulous training and evaluation, our models demonstrated promising outcomes, particularly in the nuanced domain of rice grain analysis.

The performance of the ResNet model in the task of single grain classification was quantified using precision, recall, and F1-score metrics across seven varieties of Persian rice. The model achieved an overall accuracy of 55\%, with varying degrees of success across different rice types. Notably, 'Anbar Boo (Ghareeb)' showed the highest precision (0.79) and a commendable F1-score (0.81), indicating the model's strong capability to identify this variety accurately. This high performance could be attributed to distinctive features inherent to 'Anbar Boo (Ghareeb)' that the model could efficiently learn and generalize.

Conversely, 'Sadree Dom Zard' displayed a substantial discrepancy between precision (0.71) and recall (0.08), resulting in a low F1-score of 0.15. This suggests challenges in consistently identifying 'Sadree Dom Zard' grains, possibly due to similarities with other varieties or insufficient representation in the training data. 'Ali Kazemi' and 'Sadree Dom Siahe' varieties showed remarkable recall rates of 0.61 and 0.67, respectively, though their precision scores were lower. This indicates a tendency of the model to over-classify grains as belonging to these varieties, a behavior that could lead to higher false positive rates.

The macro and weighted averages across all categories reflect a balanced model performance, with averages around 0.59 for precision and slightly lower for the F1-score, highlighting areas for improvement in the balance between model precision and recall. Table.\ref{tab:classification_results} presents the precision, recall, and F1-scores for each variety. Furthermore, Table.\ref{tab:confusion_matrix} displays the confusion matrix for the single grain classification task, offering insights into the model's performance across different rice varieties.

\begin{table}[!ht]
\centering
\begin{tabular}{|l|c|c|c|c|}
\hline
Variety & Precision & Recall & F1-score \\
\hline
Ali Kazemi & 0.43 & 0.61 & 0.51\\
Anbar Boo (Ghareeb) & 0.79 & 0.82 & 0.81 \\
Hashemi & 0.58 & 0.56 & 0.57\\
Khazar & 0.63 & 0.54 & 0.58 \\
Sadree Dom Siahe & 0.40 & 0.67 & 0.50 \\
Sadree Dom Zard & 0.71 & 0.08 & 0.15 \\
Shirodi & 0.58 & 0.52 & 0.55\\
\hline
\end{tabular}
\caption{Classification results for the Persian rice dataset.}
\label{tab:classification_results}
\end{table}

\begin{table}[!ht]
\centering
\begin{tabular}{|l|c|c|c|c|c|c|c|}
\hline
 & Ali Kazemi & Anbar Boo & Hashemi & Khazar & Sadree Dom Siahe & Sadree Dom Zard & Shirodi \\
\hline
Ali Kazemi & 36 & 3 & 5 & 4 & 11 & 0 & 0 \\
Anbar Boo & 5 & 56 & 0 & 5 & 0 & 0 & 2 \\
Hashemi & 6 & 0 & 34 & 2 & 12 & 1 & 6 \\
Khazar & 7 & 6 & 6 & 37 & 5 & 0 & 8 \\
Sadree Dom Siahe & 10 & 0 & 2 & 5 & 44 & 1 & 4 \\
Sadree Dom Zard & 9 & 4 & 9 & 1 & 27 & 5 & 4 \\
Shirodi & 10 & 2 & 3 & 5 & 11 & 0 & 33 \\
\hline
\end{tabular}
\caption{Confusion matrix for the classification of Persian rice varieties.}
\label{tab:confusion_matrix}
\end{table}


In the segmentation task, the U-Net model achieved an impressive Intersection over Union (IoU) score of 93\%, illustrating the model's efficacy in segmenting rice grains within bulk images. This high IoU score is indicative of the model's precision in distinguishing between closely packed grains and accurately delineating their boundaries, a critical step for subsequent bulk prediction analyses.

\begin{figure}[!ht]
    \centering
    \begin{subfigure}{0.48\linewidth}
        \includegraphics[width=\linewidth]{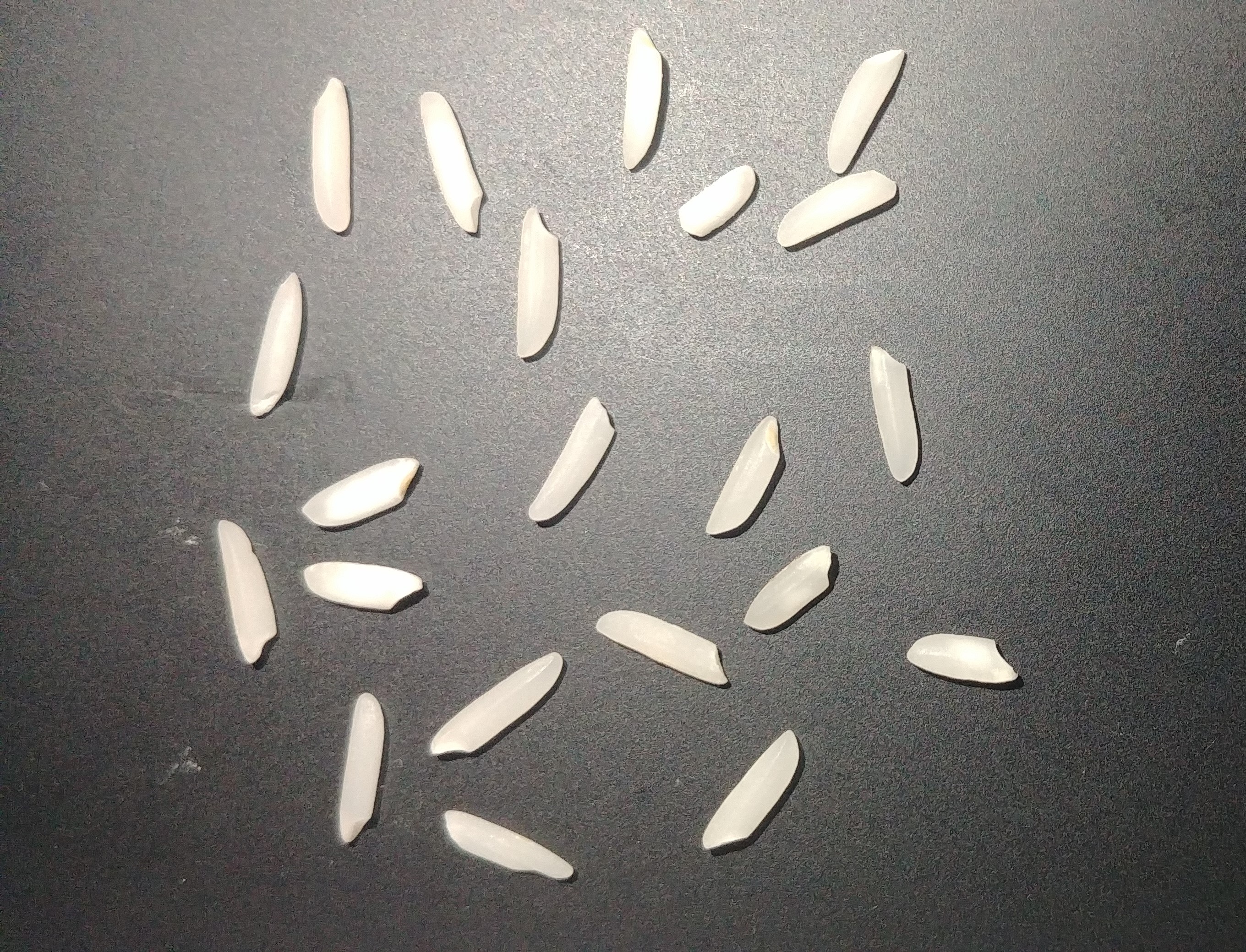}
        \caption{Sample bulk image of mixed rice varieties.}
    \end{subfigure}
    \hfill
    \begin{subfigure}{0.48\linewidth}
        \includegraphics[width=\linewidth]{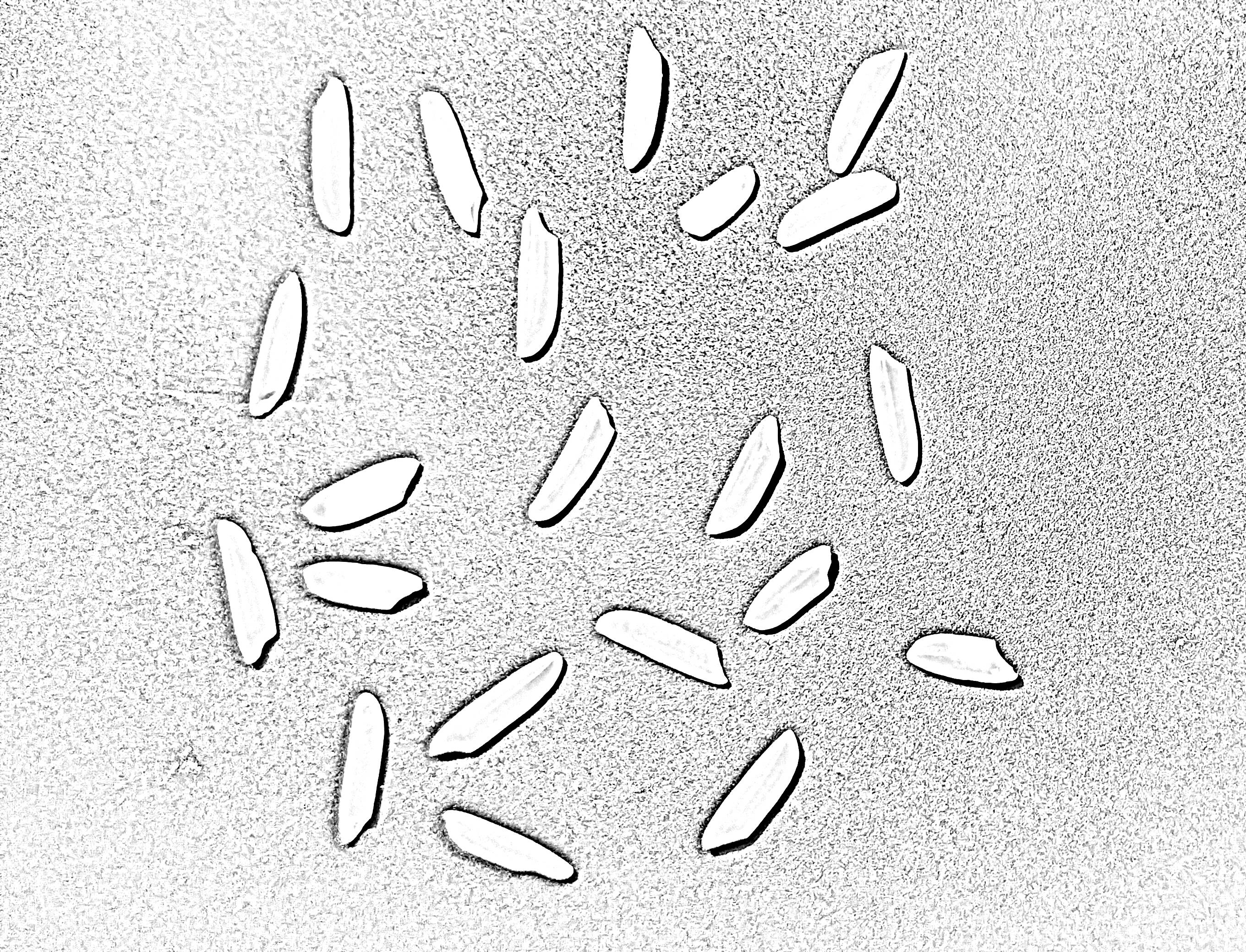}
        \caption{Segmentation result highlighting individual grains.}
    \end{subfigure}
    \caption{Comparison of a sample bulk image with its segmentation result. The segmentation demonstrates the model's ability to accurately identify and delineate individual rice grains, facilitating precise variety composition analysis.}
    \label{fig:Sample}
\end{figure}

The success of the U-Net model in segmentation underscores the potential of deep learning approaches for practical applications in the agricultural sector, especially for tasks requiring high levels of detail and accuracy in image analysis. The model's ability to generate precise segmentation masks can significantly enhance the accuracy of bulk prediction tasks, facilitating more accurate estimations of variety compositions in mixed rice samples. As illustrated in Fig.\ref{fig:Sample}, the model accurately predicts the presence of 21 rice grains in the picture, categorized into four types: it identified 7 grains as Hashemi (with 5 being the ground truth), 8 grains as Anbar Boo (5 ground truth), 3 grains as Khazar (4 ground truth), and 3 grains as Sadri Dom Siah (7 ground truth), demonstrating the model's predictive capability and its practical utility in distinguishing among various rice types within a bulk image.

\section{CONCLUSION AND FUTURE WORK}

The results underscore the nuanced capabilities of deep learning models in classifying and segmenting Persian rice varieties. While the classification task revealed areas needing refinement, particularly in balancing precision and recall, the segmentation task demonstrated the potential of using advanced deep learning techniques for detailed agricultural image analysis. The promising IoU score in segmentation tasks reinforces the value of deep learning in agricultural applications, paving the way for further exploration and implementation of these technologies in real-world settings.

Future work will focus on addressing the identified challenges in classification through more extensive data augmentation, refined model tuning, and exploring hybrid models to improve overall accuracy and consistency. A particularly promising avenue for future research is the incorporation of model explainability. This approach could offer deeper insights into the structural differences between rice varieties that the models leverage for classification and segmentation decisions. By understanding the features that models use to distinguish among varieties, researchers can refine these models for even better performance and reliability.

\printbibliography
\end{document}